\documentclass[11pt,letterpaper]{article}
\usepackage{emnlp2016}
\usepackage{times}
\usepackage{latexsym}

\emnlpfinalcopy

\def\emnlppaperid{***}

\usepackage{graphicx}
\usepackage{epstopdf}

\usepackage{multirow}
\usepackage{subfigure}
\usepackage{color}
\usepackage[table]{xcolor}
\usepackage{amssymb}
\usepackage{url}

\definecolor{0.5}{gray}{0.5}
\definecolor{0.4}{gray}{0.6}
\definecolor{0.3}{gray}{0.7}
\definecolor{0.2}{gray}{0.9}

\title{Aspect Level Sentiment Classification with Deep Memory Network}

\author{
	Duyu Tang, Bing Qin\thanks{\ \ \ Corresponding author.},\ \  Ting Liu\\
	Harbin Institute of Technology, Harbin, China\\
	\{dytang,\ qinb,\ tliu\}@ir.hit.edu.cn
}

\date{}

\begin{document}

\maketitle

\begin{abstract}
We introduce a deep memory network for aspect level sentiment classification.
Unlike feature-based SVM and sequential neural models such as LSTM, this approach explicitly captures the importance of each context word when inferring the sentiment polarity of an aspect.
Such importance degree and text representation are calculated with multiple computational layers, each of which is a neural attention model over an external memory. 
Experiments on laptop and restaurant datasets demonstrate that our approach performs comparable to state-of-art feature based SVM system,
and substantially better than LSTM and attention-based LSTM architectures.
On both datasets we show that multiple computational layers could improve the performance.
Moreover, our approach is also fast. The deep memory network with 9 layers is 15 times faster than LSTM with a CPU implementation.
\end{abstract}

\vspace{0.1cm}
\section{Introduction}
Aspect level sentiment classification is a fundamental task in the field of sentiment analysis  \cite{Pang2008,Liu2012a,Pontiki2014}.
Given a sentence and an aspect occurring in the sentence, this task aims at inferring the sentiment polarity (e.g. positive, negative, neutral) of the aspect. For example, in sentence ``\textit{great food but the service was dreadful!}'', the sentiment polarity of aspect ``\textit{food}'' is positive while the polarity of aspect ``\textit{service}'' is negative. 
Researchers typically use machine learning algorithms and build sentiment classifier in a supervised manner.
Representative approaches in literature include feature based Support Vector Machine
\cite{Kiritchenko2014-SemEval,Wagner2014-SemEval} and neural network models \cite{Dong2014,Lakkaraju2014aspect,Vo2015,Nguyen2015:EMNLP,Tang2015arxiv}.
Neural models are of growing interest for their capacity to learn text representation from data without careful engineering of features, and to capture semantic relations between aspect and context words in a more scalable way than feature based SVM.

Despite these advantages, conventional neural models like long short-term memory (LSTM) \cite{Tang2015arxiv}  capture context information in an implicit way, and are incapable of explicitly exhibiting important context clues of an aspect.
We believe that only some subset of context words are needed to infer the sentiment towards an aspect. For example, in sentence ``\textit{great food but the service was dreadful!}'', ``\textit{dreadful}'' is an important clue for the aspect ``\textit{service}'' but ``\textit{great}'' is not needed. Standard LSTM works in a sequential way and manipulates each context word with the same operation, so that it cannot explicitly reveal the importance of each context word. 
A desirable solution should be capable of explicitly capturing the importance of context words and using that information to build up features for the sentence after given an aspect word. 
%
%We believe that such aspect-specific context clues are
%crucial for a representation learning system because context words do not contribute equally with regard to an aspect. 
%For instance, in the aforementioned example, ``\textit{great}'' is a more important clue than ``\textit{dreadful}'' for the aspect ``\textit{food}''.
%On the contrary, ``\textit{dreadful}'' is more important than ``\textit{great}'' for the aspect ``\textit{service}''.
Furthermore, a human asked to do this task will selectively focus on parts of the contexts, and acquire information where it is needed to build up an internal representation towards an aspect in his/her mind.
%A desirable aspect level sentiment analyzer should be able to simulate human perception in a similar way.

In pursuit of this goal, we develop deep memory network for aspect level sentiment classification, which is inspired by the recent success of computational models with attention mechanism and explicit memory \cite{Graves2014neural,Bahdanau2015,Sukhbaatar2015end}.
Our approach is data-driven, computationally efficient and does not rely on syntactic parser or sentiment lexicon. 
The approach consists of multiple computational layers with shared parameters.
Each layer is a content- and location- based attention model, which first learns the importance/weight of each context word and then utilizes this information to calculate continuous text representation. 
The text representation in the last layer is regarded as the feature for sentiment classification. 
As every component is differentiable, the entire model could be efficiently trained end-to-end with gradient descent, where the loss function is the cross-entropy error of sentiment classification.

We apply the proposed approach to laptop and restaurant datasets from SemEval 2014 \cite{Pontiki2014}. 
Experimental results show that our approach performs comparable to a top system using feature-based SVM \cite{Kiritchenko2014-SemEval}.
On both datasets, our approach outperforms both LSTM and attention-based LSTM models \cite{Tang2015arxiv} in terms of classification accuracy and running speed. 
Lastly, we show that using multiple computational layers over external memory could achieve improved performance.

\section{Background: Memory Network}

Our approach is inspired by the recent success of memory network in question answering \cite{Weston2014memory,Sukhbaatar2015end}. 
We describe the background on memory network in this part.

Memory network is a general machine learning framework introduced by \newcite{Weston2014memory}.
Its central idea is inference with a long-term memory component, which could be read, written to, and jointly learned with the goal of using it for prediction.
Formally, a memory network consists of a memory $m$ and four components $I$, $G$, $O$ and $R$, where 
$m$ is an array of objects such as an array of vectors. 
Among these four components, $I$ converts input to internal feature representation, $G$ updates old memories with new input, $O$ generates an output representation given a new input and the current memory state, $R$ outputs a response based on the output representation. 

Let us take question answering as an example to explain the work flow of memory network. Given a list of sentences and a question, the task aims to find evidences from these sentences and generate an answer, e.g. a word.
During inference, $I$ component reads one sentence $s_i$ at a time and encodes it into a vector representation.
Then $G$ component updates a piece of memory $m_i$ based on current sentence representation. 
After all sentences are processed, we get a memory matrix $m$ which stores the semantics of these sentences, each row representing a sentence. 
Given a question $q$, memory network encodes it into vector representation $e_q$, and then $O$ component uses $e_q$ to select question related evidences from memory $m$ and generates an output vector $o$. 
Finally, $R$ component takes $o$ as the input and outputs the final response.
It is worth noting that $O$ component could consist of one or more computational layers (hops). 
The intuition of utilizing multiple hops is that more abstractive evidences could be found based on previously extracted evidences.  
\newcite{Sukhbaatar2015end} demonstrate that multiple hops could uncover more abstractive evidences than single hop, and could yield improved results on question answering and language modeling. 

\section{Deep Memory Network for Aspect Level Sentiment Classification}
In this section, we describe the deep memory network approach for aspect level sentiment classification. 
We first give the task definition. 
Afterwards, we describe an overview of the approach before presenting the content- and location- based attention models in each computational layer.
Lastly, we describe the use of this approach for aspect level sentiment classification.

\subsection{Task Definition and Notation}
Given a sentence $s=\{w_1, w_2, ..., w_i, ... w_n\}$ consisting of $n$ words and an aspect word $w_i$ \footnote{In practice, an aspect might be a multi word expression such as ``\textit{battery life}''. For simplicity we still consider aspect as a single word in this definition.} occurring in sentence $s$, aspect level sentiment classification aims at determining the sentiment polarity of sentence $s$ towards the aspect $w_i$. 
For example, the sentiment polarity of sentence ``\textit{great food but the service was dreadful!}''  towards aspect ``\textit{food}'' is positive, while the polarity towards aspect ``\textit{service}'' is negative. 
When dealing with a text corpus, we map each word into a low dimensional, continuous and real-valued vector, also known as word embedding \cite{Mikolov2013a,Pennington2014}.
All the word vectors are stacked in a word embedding matrix $L \in \mathbb{R}^{d \times |V|}$, where $d$ is the dimension of word vector and $|V|$ is vocabulary size.
The word embedding of $w_i$ is notated as $e_i \in \mathbb{R}^{d \times 1}$, which is a column in the embedding matrix $L$.

\subsection{An Overview of the Approach}
We present an overview of the deep memory network for aspect level sentiment classification.

Given a sentence $s=\{w_1, w_2, ..., w_i, ... w_n\}$ and the aspect word $w_i$, we map each word into its embedding vector. 
These word vectors are separated into two parts, aspect representation and context representation. 
If aspect is a single word like ``\textit{food}'' or ``\textit{service}'', aspect representation is the embedding of aspect word.
For the case where aspect is multi word expression like ``\textit{battery life}'', aspect representation is an average of its constituting word vectors \cite{Sun2015}. 
To simplify the interpretation, we consider aspect as a single word $w_i$. 
Context word vectors \{$e_1$, $e_2$ ... $e_{i-1}$, $e_{i+1}$  ... $e_n$\} are stacked and regarded as the external memory $m \in \mathbb{R}^{d \times (n-1)}$, where $n$ is the sentence length. 

\begin{figure}[t]
	\centering
	\includegraphics[width=.48\textwidth]{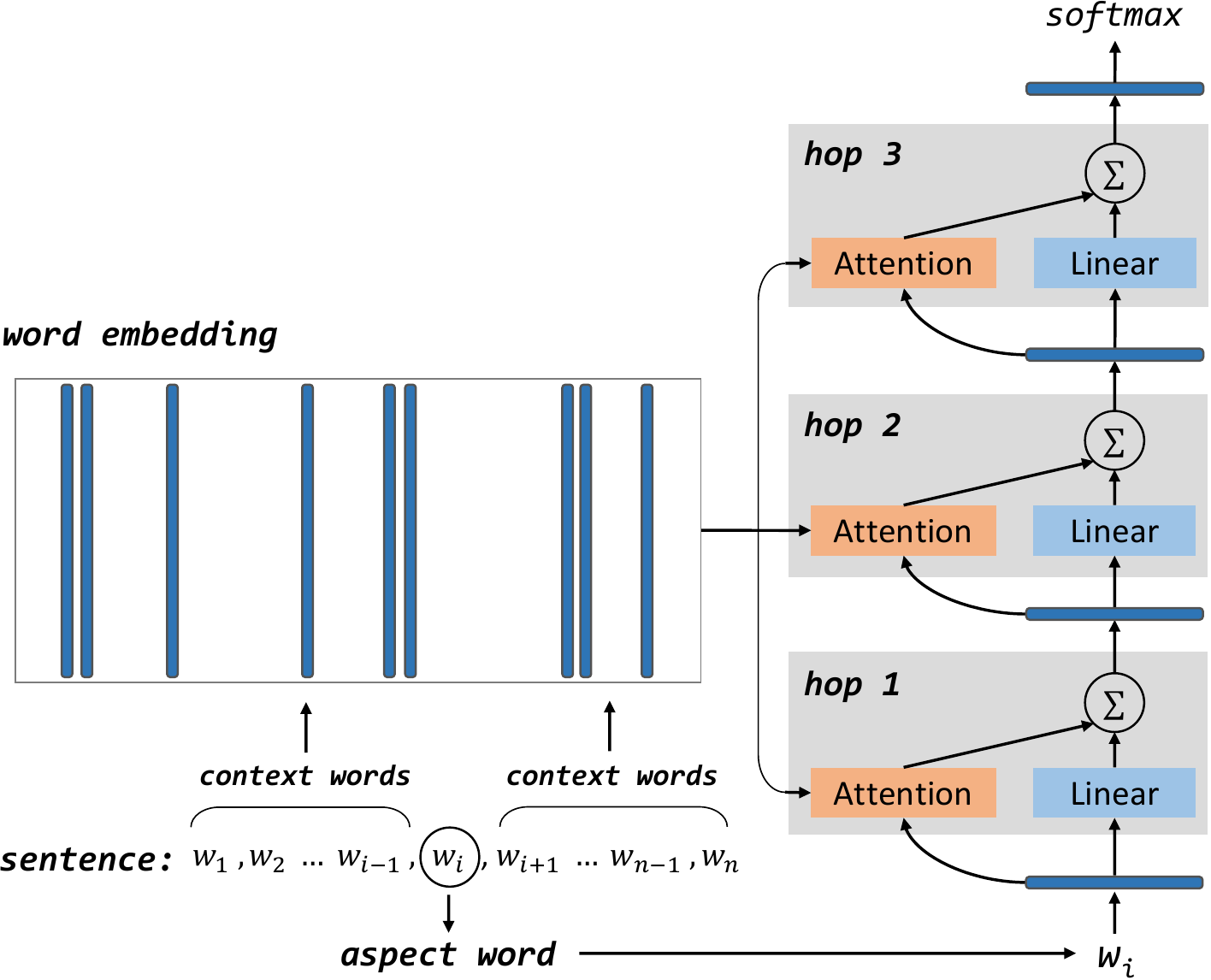}
	\caption{An illustration of our deep memory network with three computational layers (hops) for aspect level sentiment classification.}
	\label{fig:framework}
\end{figure}

An illustration of our approach is given in Figure \ref{fig:framework}, which is inspired by the use of memory network in question answering \cite{Sukhbaatar2015end}.
%Specificall
Our approach consists of multiple computational layers (hops), each of which contains an attention layer and a linear layer. 
In the first computational layer (hop 1), we regard aspect vector as the input to adaptively select important evidences from memory $m$ through attention layer. 
The output of attention layer and the linear transformation of aspect vector\footnote{In preliminary experiments, we tried directly using aspect vector without a linear transformation, and found that adding a linear layer works slightly better.} are summed and the result is considered as the input of next layer (hop 2). 
In a similar way, we stack multiple hops and run these steps multiple times, so that more abstractive evidences could be selected from the external memory $m$. 
The output vector in last hop is considered as the representation of sentence with regard to the aspect, and is further used as the feature for aspect level sentiment classification. 

It is helpful to note that the parameters of attention and linear layers are shared in different hops. 
%We describe the details of attention models in following subsections.
Therefore, the model with one layer and the model with nine layers have the same number of parameters.

\subsection{Content Attention}
We describe our attention model in this part.
The basic idea of attention mechanism is that it assigns a weight/importance to each lower position when computing an upper level representation \cite{Bahdanau2015}.
In this work, we use attention model to compute the representation of a sentence with regard to an aspect. 
The intuition is that context words do not contribute equally to the semantic meaning of a sentence. 
Furthermore, the importance of a word should be different if we focus on different aspect.
Let us again take the example of  ``\textit{great food but the service was dreadful!}''. 
The context word ``\textit{great}'' is more important than ``\textit{dreadful}'' for aspect ``\textit{food}''.
On the contrary, ``\textit{dreadful}'' is more important than ``\textit{great}'' for aspect ``\textit{service}''.

Taking an external memory $m \in \mathbb{R}^{d \times k}$ and an aspect vector $v_{aspect} \in \mathbb{R}^{d \times 1}$ as input, the attention model outputs a continuous vector $vec \in \mathbb{R}^{d \times 1}$.
The output vector is computed as a weighted sum of each piece of memory in $m$, namely 
\begin{equation}
vec = \sum_{i=1}^{k}\alpha_i m_i
\end{equation}
where $k$ is the memory size, $\alpha_i \in [0,1]$ is the weight of $m_i$ and $\sum_{i} \alpha_i = 1$.
We implement a neural network based attention model.
%, whose scoring architecture is shown in Figure~\ref{fig:attention}.
%\begin{figure}[h]
%	\centering
%	\includegraphics[width=.45\textwidth]{fig-attention-2.pdf}
%	\caption{Our neural attention architecture.}
%	\label{fig:attention}
%\end{figure}
For each piece of memory $m_i$, we use a feed forward neural network to compute its semantic relatedness with the aspect. The scoring function is calculated as follows, where $W_{att} \in \mathbb{R}^{1 \times 2d}$ and $b_{att} \in \mathbb{R}^{1 \times 1}$.
\begin{equation}
	g_i = tanh(W_{att} [m_i; v_{aspect}] + b_{att})
\end{equation}
After obtaining \{$g_1$, $g_2$, ... $g_k$\}, we feed them to a $softmax$ function to calculate the final importance scores \{$\alpha_1$, $\alpha_2$, ... $\alpha_k$\}.
%, namely $\alpha_i =exp(g_i) / \sum_{j=1}^k exp(g_{j})$.
\begin{equation}
	\alpha_i = \frac{exp(g_i)}{\sum_{j=1}^k exp(g_{j})}
\end{equation}

We believe that such an attention model has two advantages. 
One advantage is that this model could adaptively assign an importance score to each piece of memory $m_i$ according to its semantic relatedness with the aspect.
Another advantage is that this attention model is differentiable, so that it could be easily trained together with other components in an end-to-end fashion. 

\subsection{Location Attention}\label{section:location-attention}
We have described our neural attention framework and a content-based model in previous subsection. 
However, the model mentioned above 
ignores the location information between context word and aspect. 
Such location information is helpful for an attention model because intuitively a context word closer to the aspect should be more important than a farther one.
In this work, we define the location of a context word as its absolute distance with the aspect in the original sentence sequence\footnote{The location of a context word could also be measured by its distance to the aspect along a syntactic path. We leave this as a future work as we prefer to developing a purely data-driven approach without using external parsing results.}.
On this basis, we study four strategies to encode the location information in the attention model.
The details are described below.

$\bullet$ {Model 1}. Following \newcite{Sukhbaatar2015end}, we calculate the memory vector $m_i$ with 
\begin{equation}
m_i = e_i \odot v_{i}
\end{equation}
where $\odot$ means element-wise multiplication and $v_{i} \in \mathbb{R}^{d \times 1}$ is a location vector for word $w_i$. 
Every element in $v_i$ is calculated as follows,
\begin{equation}
	v_{i}^{k} = (1 - l_i/n) - (k/d) (1 - 2 \times l_i/n)
\end{equation}
where $n$ is sentence length,  $k$ is the hop number and $l_i$ is the location of $w_i$. 
%The location score of each word 

$\bullet$ {Model 2}. This is a simplified version of Model 1, using the same location vector $v_i$ for $w_i$ in different hops.
Location vector $v_i$ is calculated as follows.
% $v_{i} = 1 - l_i/n$.
\begin{equation}
v_{i} = 1 - l_i/n
\end{equation}

$\bullet$ {Model 3}. We regard location vector $v_i$ as a parameter and compute a piece of memory with vector addition, namely 
\begin{equation}
m_i = e_i + v_i
\end{equation}

All the position vectors are stacked in a position embedding matrix, which is jointly learned with gradient descent.

$\bullet$ {Model 4}. Location vectors are also regarded as parameters. 
Different from Model 3, location representations are regarded as neural gates to control how many percent of word semantics is written into the memory. We feed location vector $v_i$ to a sigmoid function $\sigma$, and calculate $m_i$ with element-wise multiplication: 
\begin{equation}
m_i = e_i \odot \sigma(v_i)
\end{equation}

\subsection{The Need for Multiple Hops}
It is widely accepted that computational models that are composed of multiple processing layers have the ability to learn representations of data with multiple levels of abstraction \cite{LeCun2015}.
In this work, the attention layer in one layer is essentially a weighted average compositional function, which is not powerful enough to handle the sophisticated computationality like negation, intensification and contrary in language.
Multiple computational layers allow the deep memory network to learn representations of text with multiple levels of abstraction. Each layer/hop retrieves important context words, and transforms the representation at previous level into a representation at a higher, slightly more abstract level. With the composition of enough such transformations, very complex functions of sentence representation towards an aspect can be learned.

\subsection{Aspect Level Sentiment Classification}
We regard the output vector in last hop as the feature, and feed it to a $softmax$ layer for aspect level sentiment classification. 
The model is trained in a supervised manner by minimizing the cross entropy error of sentiment classification, whose loss function is given below, where $T$ means all training instances, $C$ is the collection of sentiment categories, $(s, a)$ means a sentence-aspect pair.

%The parameters in multiple hops are shared. 
\begin{equation}
	loss = -\sum_{(s,a) \in T}^{}\sum_{c \in C} P_{c}^{g}(s, a) \cdot log(P_{c}(s, a))
\end{equation}
$P_c(s, a)$ is the probability of predicting $(s, a)$ as category $c$ produced by our system. 
$P^g_c(s, a)$ is 1 or 0, indicating whether the correct answer is $c$.
We use back propagation to calculate the gradients of all the parameters, 
%calculate the  derivative of loss function through back-propagation with respect to all parameters, 
and update them with stochastic gradient descent. 
We clamp the word embeddings with 300-dimensional Glove vectors \cite{Pennington2014}, which is trained from web data and the vocabulary size is 1.9M\footnote{Available at: {http://nlp.stanford.edu/projects/glove/}.}.
We randomize other parameters with uniform distribution $U(-0.01, 0.01)$, and set the learning rate as 0.01.

%\begin{figure*}[t]
%	\centering
%	\includegraphics[width=\textwidth]{ppt1.pdf}
%	\caption{A deep memory network for aspect level sentiment classification.}
%	\label{fig:framework}
%\end{figure*}

\section{Experiment}
We describe experimental settings and report empirical results in this section.

\subsection{Experimental Setting}
We conduct experiments on two datasets from SemEval 2014 \cite{Pontiki2014}, one from laptop domain and another from restaurant domain. 
Statistics of the datasets are given in Table \ref{table:dataset}. {It is worth noting that the original dataset contains the fourth category - conflict, which means that a sentence expresses both positive and negative opinion towards an aspect. We remove conflict category as the number of instances is very tiny, incorporating which will make the dataset extremely unbalanced.} Evaluation metric is classification accuracy. 
\begin{table}[h]
	\centering
	\begin{tabular}{l|c|c|c}
		\hline
		Dataset & Pos. & Neg. & Neu. \\
		\hline
		Laptop-Train & 994 & 870 & 464\\
		Laptop-Test & 341 & 128 & 169\\
		Restaurant-Train & 2164& 807 & 637\\
		Restaurant-Test & 728& 196& 196\\
		\hline
	\end{tabular}
	\caption{Statistics of the datasets.}
	\label{table:dataset}
\end{table}

%\cite{Manning2014}
%fdf

%\begin{table}[h]
%	\centering
%	\begin{tabular}{l|c|c|c|c}
%		\hline
%		Dataset & Pos. & Neg. & Neu. & Con. \\
%		\hline
%		Laptop-Train & 994 & 870 & 464& 45\\
%		Laptop-Test & 341 & 128 & 169& 16\\
%		Restaurant-Train & 2164& 807 & 637& 91\\
%		Restaurant-Test & 728& 196& 196& 14\\
%		\hline
%	\end{tabular}
%	\label{table:dataset}
%	\caption{Aspect terms and their polarities per domain.  }
%\end{table}

\subsection{Comparison to Other Methods}
We compare with the following baseline methods on both datasets.

(1) \textbf{Majority} is a basic baseline method, which assigns the majority sentiment label in training set to each instance in the test set.

(2) \textbf{Feature-based SVM} performs state-of-the-art on aspect level sentiment classification.
We compare with a top system using ngram features, parse features and lexicon features \cite{Kiritchenko2014-SemEval}. 
%The sentiment lexicons are from \cite{Hu2004,Wilson2005,Mohammad2012,Mohammad2013}.

(3) We compare with three LSTM models \cite{Tang2015arxiv}). In \textbf{LSTM}, a LSTM based recurrent model is applied from the start to the end of a sentence, and the last hidden vector is used as the sentence representation. \textbf{TDLSTM} extends LSTM by taking into account of the aspect, and uses two LSTM networks, a forward one and a backward one, towards the aspect.
\textbf{TDLSTM+ATT} extends TDLSTM by incorporating an attention mechanism \cite{Bahdanau2015} over the  hidden vectors.
We use the same Glove word vectors for fair comparison. 
%For further details see \cite{Tang2015arxiv}.

(4) We also implement \textbf{ContextAVG}, a simplistic version of our approach. Context word vectors are averaged and the result is added to the aspect vector. 
%The final text representation is fed to $softmax$ for sentiment classification. 
The output is fed to a $softmax$ function.

\begin{table}[h]
	\centering
	\begin{tabular}{l|c|c}
		\hline
		& {Laptop} & {Restaurant} \\
		\hline
		Majority 				& 53.45	& 65.00 \\
		Feature+SVM				& \textbf{72.10}	& \textbf{80.89} \\
		LSTM					& 66.45 & 74.28 \\
		TDLSTM					& 68.13 & 75.63 \\
		TDLSTM+ATT				& 66.24 & 74.31 \\
		ContextAVG				& 61.22	& 71.33	\\
		\hline
		MemNet (1)				& 67.66 & 76.10 \\
		MemNet (2)				& 71.14 & 78.61 \\
		MemNet (3)				& 71.74 & 79.06 \\
		MemNet (4)				& 72.21 & 79.87 \\
		MemNet (5)				& 71.89 & 80.14 \\
		MemNet (6)				& 72.21 & 80.05 \\
		MemNet (7)				& \textbf{72.37} & 80.32 \\
		MemNet (8)				& 72.05 & 80.14 \\
		MemNet (9)				& 72.21 & \textbf{80.95} \\
		\hline
	\end{tabular}	
	\caption{Classification accuracy of different methods on laptop and restaurant datasets. Best scores in each group are in bold.}
	\label{table:experiment-baseline}
\end{table}

Experimental results are given in Table \ref{table:experiment-baseline}. 
Our approach using only content attention is abbreviated to MemNet ($k$), where $k$ is the number of hops. 
%It is unsurprisingly that majority performs the worse among all these baseline methods. 
%From Table \ref{table:experiment-baseline}, w
We can find that feature-based SVM is an extremely strong performer and substantially outperforms other baseline methods, which demonstrates the importance of a powerful feature representation for aspect level sentiment classification. 
Among three recurrent models, TDLSTM performs better than LSTM, which indicates that taking into account of the aspect information is helpful. 
This is reasonable as the sentiment polarity of a sentence towards different aspects (e.g. ``\textit{food}'' and ``\textit{service}'') might be different. 
It is somewhat disappointing that incorporating attention model over TDLSTM does not bring any improvement. 
We consider that each hidden vector of TDLSTM encodes the semantics of word sequence until the current position. Therefore, the model of TDLSTM+ATT actually selects such mixed semantics of word sequence, which is weird and not an intuitive way to selectively focus on parts of contexts. 
Different from TDLSTM+ATT, the proposed memory network approach removes the recurrent calculator over word sequence and directly apply attention mechanism on context word representations.

We can also find that the performance of ContextAVG is very poor, which means that assigning the same weight/importance to all the context words is not an effective way. 
Among all our models from single hop to nine hops, we can observe that using more computational layers could generally lead to better performance, especially when the number of hops is less than six.
The best performances are achieved when the model contains seven and nine hops, respectively. 
On both datasets, the proposed approach could obtain comparable accuracy compared to the state-of-art feature-based SVM system.

\subsection{Runtime Analysis}
We study the runtime of recurrent neural models and the proposed deep memory network approach with different hops. 
We implement all these approaches based on the same neural network infrastructure, use the same 300-dimensional Glove word vectors, and run them on the same CPU server. 

%because the recurrent calculators of sequence length are removed. 
\begin{table}[h]
	\centering
	\begin{tabular}{l|c}
		\hline
		Method & Time cost\\
		\hline
		LSTM & 417 \\
		TDLSTM & 490 \\
		TDLSTM + ATT & 520 \\
		\hline
		MemNet (1) & 3 \\
		MemNet (2)& 7\\
		MemNet (3)& 9 \\
		MemNet (4)& 15\\
		MemNet (5)& 20\\
		MemNet (6)& 24\\
		MemNet (7)& 26\\
		MemNet (8)& 27\\
		MemNet (9)& 29\\
		\hline
	\end{tabular}
	\caption{Runtime (seconds) of each training epoch on the restaurant dataset.}
	\label{table:time-cost}
\end{table}
The training time of each iteration on the restaurant dataset is given in Table \ref{table:time-cost}. 
We can find that LSTM based recurrent models are indeed computationally expensive, which is caused by the complex operations in each LSTM unit along the word sequence. 
Instead, the memory network approach is simpler and evidently faster because it does not need recurrent calculators of sequence length.
Our approach with nine hops is almost 15 times faster than the basic LSTM model. 

\subsection{Effects of Location Attention}
As described in Section \ref{section:location-attention}, we explore four strategies to integrate location information into the attention model. 
We incorporate each of them separately into the basic content-based attention model.
It is helpful to restate that the difference between four location-based attention models lies in the usage of location vectors for context words. In Model 1 and Model 2, the values of location vectors are fixed and calculated in a heuristic way. In Model 3 and Model 4,
location vectors are also regarded as the parameters and jointly learned along with other parameters in the deep memory network. 

\begin{figure}[h]
	\centering
	\includegraphics[width=.48\textwidth]{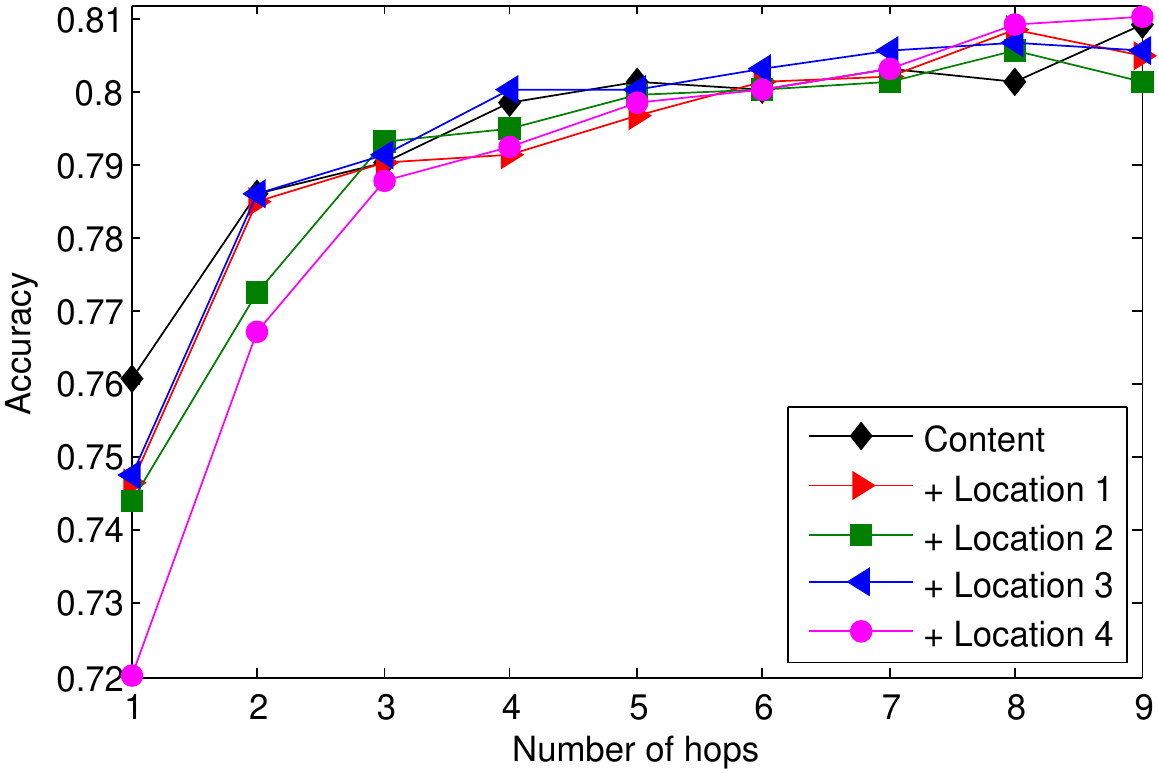}
	\caption{Classification accuracy of different attention models on the restaurant dataset. }
	\label{fig:location}
\end{figure}

\begin{table*}[t]\footnotesize
	\centering
	\subtable[Aspect: {\textit{service}}, Answer: -1, Prediction: -1]{
		\begin{tabular}{c|c|c|c|c|c}
			\hline
			& hop 1 & hop 2 & hop 3 & hop 4 & hop 5 \\
			\hline
			great & \cellcolor{0.2}0.20  & 0.15  & 0.14  & 0.13  & \cellcolor{0.2}0.23  \\
			food & 0.11  & 0.07  & 0.08  & 0.12  & 0.06  \\
			but & \cellcolor{0.2}0.20  & 0.10  & 0.10  & 0.12  & 0.13  \\
			the & 0.03  & 0.07  & 0.08  & 0.12  & 0.06  \\
			was & 0.08  & 0.07  & 0.08  & 0.12  & 0.06  \\
			dreadful & \cellcolor{0.2}0.20  & \cellcolor{0.4}0.45  & \cellcolor{0.4}0.45  & \cellcolor{0.3}0.28  & \cellcolor{0.4}0.40  \\
			! & 0.19  & 0.08  & 0.08  & 0.12  & 0.07  \\
			\hline
		\end{tabular}
	}\ \ \ \ \ \ \ \ 
	\subtable[Aspect: {\textit{food}}, Answer: +1, Prediction: -1]{
		\begin{tabular}{c|c|c|c|c|c}
			\hline
			& hop 1 & hop 2 & hop 3 & hop 4 & hop 5 \\
			\hline
			great & \cellcolor{0.2}0.22  & 0.12  & 0.14  & 0.12  & \cellcolor{0.2}0.20  \\
			but & \cellcolor{0.2}0.21  & 0.11  & 0.10  & 0.11  & 0.12  \\
			the & 0.03  & 0.11  & 0.08  & 0.11  & 0.06  \\
			service & 0.11  & 0.11  & 0.08  & 0.11  & 0.06  \\
			was & 0.04  & 0.11  & 0.08  & 0.11  & 0.06  \\
			dreadful & \cellcolor{0.2}0.22  & \cellcolor{0.3}0.32  & \cellcolor{0.4}0.45  & \cellcolor{0.3}0.32  & \cellcolor{0.4}0.43  \\
			! & 0.16  & 0.11  & 0.08  & 0.11  & 0.07  \\
			\hline
		\end{tabular}
	}
	\caption{Examples of attention weights in different hops for aspect level sentiment classification. The model only uses content attention. The hop columns show the weights of context words in each hop, indicated by values and gray color. This example shows the results of sentence ``\textit{great food but the service was dreadful!}'' with ``\textit{food}'' and ``\textit{service}'' as the aspects.}
	\label{table:case-only-context}
	%Restaurant-hop9-model1--12.model.attentionResults
\end{table*}

\begin{table*}[t]\footnotesize
	\centering
	\subtable[Aspect: {\textit{service}}, Answer: -1, Prediction: -1]{
		\begin{tabular}{c|c|c|c|c|c}
			\hline
			& hop 1 & hop 2 & hop 3 & hop 4 & hop 5 \\
			\hline
			great & 0.08  & 0.10  & 0.10  & 0.09  & 0.09 \\
			food & 0.08  & 0.07  & 0.07  & 0.07  & 0.07 \\
			but & 0.10  & 0.15  & 0.16  & 0.13  & 0.11 \\
			the & 0.07  & 0.07  & 0.07  & 0.07  & 0.07 \\
			was & 0.07  & 0.07  & 0.07  & 0.07  & 0.07 \\
			dreadful & \cellcolor{0.5}0.52  & \cellcolor{0.4}0.48  & \cellcolor{0.4}0.48  & \cellcolor{0.5}0.50  & \cellcolor{0.5}0.52 \\
			! & 0.07  & 0.07  & 0.07  & 0.07  & 0.07 \\
			\hline
		\end{tabular}
	}\ \ \ \ \ \ \ \ 
	%Restaurant-hop6-position-2-model2--32.model.attentionResults
	\subtable[Aspect: {\textit{food}}, Answer: +1, Prediction: +1]{
		\begin{tabular}{c|c|c|c|c|c}
			\hline
			& hop 1 & hop 2 & hop 3 & hop 4 & hop 5 \\
			\hline
			great & \cellcolor{0.3}0.31 & \cellcolor{0.3}0.26 & \cellcolor{0.3}0.32 & \cellcolor{0.3}0.28 & \cellcolor{0.3}0.32 \\
			but & 0.14 & 0.18 & 0.15 & 0.18 & 0.15 \\
			the & 0.08 & 0.05 & 0.08 & 0.05 & 0.07 \\
			service & 0.09 & 0.09 & 0.09 & 0.08 & 0.09 \\
			was & 0.09 & 0.08 & 0.09 & 0.08 & 0.08 \\
			dreadful & 0.18 & \cellcolor{0.2}0.21 & 0.18 & \cellcolor{0.2}0.22 & 0.19 \\
			! & 0.11 & 0.12 & 0.10 & 0.11 & 0.10 \\
			\hline
		\end{tabular}
	}
	%Restaurant-hop7-position-2-model12--5.model.attentionResults
	\caption{Examples of attention weights in different hops for aspect level sentiment classification. The model also takes into account of the location information (Model 2). This example is as same as the one we use in Table \ref{table:case-only-context}.}
	\label{table:case-context-plus-position}
\end{table*}
Figure \ref{fig:location} shows the classification accuracy of each attention model on the restaurant dataset. 
We can find that using multiple computational layers could consistently improve the classification accuracy in all these models. 
%Incorporating location information does not have a positive impact on the performance when the number of hops is less than three.
%The reason might be that the content-based attention model is not powerful enough in this situation. 
%Mixing with more parameters like location embedding in Model 4 makes it harder to optimize the entire model.
All these models perform comparably when the number of hops is larger than five. 
Among these four location-based models, we prefer Model 2 as it is intuitive and has less computation cost without loss of accuracy. 
We also find that Model 4 is very sensitive to the choice of neural gate. 
Its classification accuracy decreases by almost 5 percentage when the $sigmoid$ operation over location vector is removed.

%\vspace{-0.5cm}
\subsection{Visualize Attention Models}
We visualize the attention weight of each context word to get a better understanding of the deep memory network approach.
The results of context-based model and location-based model (Model 2) are given in Table \ref{table:case-only-context} and Table \ref{table:case-context-plus-position}, respectively.

From Table \ref{table:case-only-context}(a), we can find that in the first hop the context words ``\textit{great}'', ``\textit{but}'' and ``\textit{dreadful}'' contribute equally to the aspect ``\textit{service}''. 
While after the second hop, the weight of ``\textit{dreadful}'' increases and finally the model correctly predict the polarity towards ``\textit{service}'' as negative. 
This case shows the effects of multiple hops.
However, in Table \ref{table:case-only-context}(b), the content-based model also gives a larger weight to ``\textit{dreadful}'' when the target we focus on is ``\textit{food}''. 
As a result, the model incorrectly predicts the polarity towards ``\textit{food}'' as negative.
This phenomenon might be caused by the neglect of location information. 
From Table \ref{table:case-context-plus-position}(b), we can find that the weight of ``\textit{great}'' is increased when the location of context word is considered. 
Accordingly, Model 2 predicts the correct sentiment label towards ``\textit{food}''.
We believe that location-enhanced model captures both content and location information. 
For instance, in Table \ref{table:case-context-plus-position}(a) the closest context words of the aspect ``\textit{service}'' are ``\textit{the}'' and ``\textit{was}'', while ``\textit{dreadful}'' has the largest weight.

\subsection{Error Analysis}
We carry out an error analysis of our location enhanced model (Model 2) on the restaurant dataset, and find that most of the errors could be summarized as follows.
The first factor is non-compositional sentiment expression. This model regards single context word as the basic computational unit and cannot handle this situation.
An example is ``\textit{\underline{dessert} was also to die for!}'', where the aspect is underlined.
The sentiment expression is ``\textit{die for}'', whose meaning could not be composed from its constituents ``\textit{die}'' and ``\textit{for}''.
% as the model regards every single context word as the basic computational unit.
%An example is ``\textit{\underline{dessert} was also to die for!}'', where the aspect is underlined.
%The sentiment expression in this example is ``\textit{die for}'', whose meaning could not be composed with the constituents ``\textit{die}'' and ``\textit{for}''.
%Second (remove this), there are cases that 
%\textit{while there 's a decent \underline{menu}, it should n't take ten minutes to get your drinks and 45 for a dessert pizza.}
%targetText: menu
The second factor is complex aspect expression consisting of many words, such as 
%``\textit{\underline{new hamburger with special sauce} is ok - at least better than big mac!}'' and
``\textit{ask for the \underline{round corner table next to the large window}.}''
This model represents an aspect expression by averaging its constituting word vectors, which could not well handle this situation.
The third factor is sentimental relation between context words such as negation, comparison and condition.
%Examples include 
%``\textit{if you are a \underline{tequila} fan you will not be disappointed}'' and 
An example is ``\textit{but \underline{dinner} here is never disappointing, even if the prices are a bit over the top}''.
We believe that this is caused by the weakness of weighted average compositional function in each hop.
There are also cases when comparative opinions are expressed such as ``\textit{i 've had better \underline{japanese food} at a mall food court}''.

\section{Related Work}
This work is connected to three research areas in natural language processing. 
We briefly describe related studies in each area. 

\subsection{Aspect Level Sentiment Classification}
Aspect level sentiment classification is a fine-grained classification task in sentiment analysis, which
%Different from course-grained sentiment classification that determines the general polarity of a sentence without considering the aspect, this task
aims at identifying the sentiment polarity of a sentence expressed towards an aspect \cite{Pontiki2014}. 
Most existing works use machine learning algorithms, and build sentiment classifier from sentences with manually annotated polarity labels.
One of the most successful approaches in literature is feature based SVM. Experts could design effective feature templates and make use of external resources like parser and sentiment lexicons \cite{Kiritchenko2014-SemEval,Wagner2014-SemEval}.
In recent years, neural network approaches \cite{Dong2014,Lakkaraju2014aspect,Nguyen2015:EMNLP,Tang2015arxiv} are of growing attention for their capacity to learn powerful text representation from data. 
However, these neural models (e.g. LSTM) are computationally expensive, and could not explicitly reveal the importance of context evidences with regard to an aspect. 
Instead, we develop simple and fast approach that explicitly encodes the context importance towards a given aspect.
It is worth noting that the task we focus on differs from fine-grained opinion extraction, which assigns each word a tag (e.g. B,I,O) to indicate whether it is an aspect/sentiment word \cite{Choi2010,Irsoy2014,Liu2015}. The aspect word in this work is given as a part of the input.
%, and the goal is to infer the sentiment polarity of a sentence towards the aspect.

\subsection{Compositionality in Vector Space}
In NLP community, compositionality means that the meaning of a composed expression (e.g. a phrase/sentence/document) comes from the meanings of its constituents \cite{Frege1892}.
\newcite{Mitchell2010} exploits a variety of addition and multiplication functions to calculate phrase vector. 
\newcite{Yessenalina2011} use matrix multiplication as compositional function to compute vectors for longer phrases.
To compute sentence representation, researchers develop denoising autoencoder \cite{Glorot2011}, convolutional neural network \cite{Kalchbrenner2014,Kim2014,Yin-schutze:2015:CoNLL}, sequence based recurrent neural models \cite{Sutskever2014sequence,Kiros2015skip,Li2015a} and tree-structured neural networks \cite{Socher2013a,Tai2015,Zhu2015}.
Several recent studies calculate continuous representation for documents with neural networks \cite{Le2014,Bhatia2015,Li2015,Tang2015,Yang2016hierarchical}.

\subsection{Attention and Memory Networks}
Recently, there is a resurgence in computational models with attention mechanism and explicit memory to learn representations of texts \cite{Graves2014neural,Weston2014memory,Sukhbaatar2015end,Bahdanau2015}. 
In this line of research, memory is encoded as a continuous representation and operations on memory (e.g. reading and writing) are typically implemented with neural networks. 
Attention mechanism could be viewed as a compositional function, where lower level representations are regarded as the memory, and the function is to choose ``where to look'' by assigning a weight/importance to each lower position when computing an upper level representation.
Such attention based approaches have achieved promising performances on a variety of NLP tasks \cite{Luong2015EMNLP,Kumar2015ask,Rush2015}.
%, including machine translation \cite{Luong2015EMNLP}, conversational agents \cite{Shang2015}, question answering \cite{Kumar2015ask}, syntactic parsing \cite{Vinyals2015grammar}, language modeling \cite{Tran2016recurrent}, semantic parsing \cite{Dong2016language}, text entailment \cite{Rocktaschel2015reasoning}, abstractive summarization \cite{Rush2015} and reading comprehension \cite{Hermann2015teaching}.

%When manipulating a memory, one could use single or multiple computational steps. \newcite{Sukhbaatar2015end} show that using multiple computational operators is crucial to achieving a good performance on question answering and language modeling. In this work, we closely follow \newcite{Sukhbaatar2015end} and develop effective deep memory networks to deal with aspect level sentiment classification. 

\section{Conclusion}
We develop deep memory networks that capture importances of context words for aspect level sentiment classification.
Compared with recurrent neural models like LSTM, this approach is simpler and faster.
Empirical results on two datasets verify that the proposed approach performs comparable to state-of-the-art feature based SVM system, and substantively better than LSTM architectures.
We implement different attention strategies
%strategies to capture the relation between context word and the aspect,
and show that leveraging both content and location information could learn better context weight and text representation. 
We also demonstrate that using multiple computational layers in memory network could obtain improved performance.
Our potential future plans are incorporating sentence structure like parsing results into the deep memory network.

\section*{Acknowledgments}
We would especially want to thank Xiaodan Zhu  for running their system on our setup.
We greatly thank Yaming Sun for tremendously helpful discussions. 
We also thank the anonymous reviewers for their valuable comments.
This work was supported by the National High Technology Development 863 Program of China (No. 2015AA015407), National Natural Science Foundation of China (No. 61632011 and No.61273321).
%According to the meaning given to this role by Harbin Institute of Technology, the contact author of this paper is Bing Qin.

\bibliography{bibtex}
\bibliographystyle{emnlp2016}

\end{document}